# Probabilistic Assumption-Based Reasoning

**Jürg Kohlas**
University of Fribourg
Institute for Informatics
CH-1700 Fribourg
Switzerland
E-mail kohlas@cfruni51

**Paul-André Monney**
University of Fribourg
Institute for Informatics
CH-1700 Fribourg
Switzerland
E-mail monney@cfruni51

## Abstract

The classical propositional assumption-based model is extended to incorporate probabilities for the assumptions. Then it is placed into the framework of evidence theory. Several authors like Laskey, Lehner (1989) and Provan (1990) already proposed a similar point of view, but the first paper is not as much concerned with mathematical foundations, and Provan's paper develops into a different direction. Here we thoroughly develop and present the mathematical foundations of this theory, together with computational methods adapted from Reiter, De Kleer (1987) and Inoue (1992). Finally, recently proposed techniques for computing degrees of support are presented.

## 1 MODELING PROPOSITIONAL SYSTEMS

The following simple example illustrates the kind of models considered in this paper. Suppose that a burglar alarm system is installed in a house. Being an electronic system there is some risk that the system fails to operate properly. In particular, the system may not react to a burglary because it is not functioning correctly. Let $a_1$ denote the event that the system is functioning correctly. The probability of $a_1$ measures the reliability of the burglar alarm system and let $q_1 = 0.95$ denote this probability. An alarm system may however respond to other stimuli than a burglar, for example to strong vibrations as generated by earthquakes. Denote by $a_2$ the presence of such other causes for an alarm and suppose its probability is $q_2 = 0.01$. This probability measures the over-sensitivity of the system. Moreover, we suppose that $a_1$ and $a_2$ are independent. The causal mechanism of this burglar alarm system can then be described by the following rules, which can of course be transformed into clauses:

$$R_1: \quad \text{if burglary} \wedge a_1 \text{ then alarm} \tag{1}$$

$$R_2: \quad \text{if } a_2 \wedge a_1 \text{ then alarm} \tag{2}$$

$$R_3: \quad \text{if } \neg \text{ burglary} \wedge \neg a_2 \text{ then } \neg \text{ alarm.} \tag{3}$$

Now suppose that the alarm sounds. Under the assumption that $\neg a_2$ is true, this fact permits to prove burglary. Since $\neg a_2$ is true with probability 1-0.01, we say that the credibility of burglary is 0.99 or also that the hypothesis of a burglary is supported to the degree 0.99 by the available information. The following is a formalization of this kind of model and of its analysis. Let $P = \{p_1, p_2, \ldots, p_n\}$ and $A = \{a_1, a_2, \ldots, a_s\}$ be two sets of propositions, the elements of the latter being called assumptions, so that $N = P \cup A$ is the entire set of symbols considered. We assume that an assumption $a_i$ is true with a known probability $q_i$ and that the assumptions are stochastically independent. If $\Sigma$ is a set of logical formulas over the alphabet $N$, then $\sim \Sigma$ denotes the set obtained by taking the negation of each formula in $\Sigma$. Thus, $P^{\pm} = P \cup \sim P$ and $A^{\pm} = A \cup \sim A$ are the sets of all literals over $P$ and $A$ respectively. Let $\Sigma_K$ denote a set of clauses composed of literals in $N^{\pm} = N \cup \sim N$ describing the available general knowledge about the situation or problem under investigation. This knowledge is uncertain because we are not sure whether the assumptions hold or not. In a given situation, further observed facts represented by clauses forming a set $\Sigma_F$ may be added to $\Sigma_K$. The clauses in $\Sigma = \Sigma_K \cup \Sigma_F$ thus represent all the available information at a given moment. These clauses are statements which restrict the common truth values of the propositions involved. Let $h$ be a formula in the language $\mathcal{L}_N$ composed of all formulas which can be formed with propositions in $N$. The formula $h$ represents a hypothesis which can be true or false, and the problem considered here is to judge how credible is $h$ in view of the available knowledge. This can be done by searching for the assumptions whose validity, together with $\Sigma$, will permit to infer $h$ and then to weight them according to their probabilities. Let $f$ be an arbitrary formula in $\mathcal{L}_N$. Each possible interpretation of the formula is then given by an $n$-tuple $x$ in the Boolean cube $B_{n+s} = \{0,1\}^{n+s}$, 1 corresponding to true and 0 to false. If $f(x)$ denotes the truth value of $f$ under the interpretation $x \in B_{n+s}$, then $f$ becomes an application from $B_{n+s}$ to $\{0,1\}$. This permits to

define
$$N(f) = \{x \in B_{n+s} : f(x) = 1\}. \quad (4)$$

This puts a bridge between propositional calculus and the algebra of sets, a link which becomes especially important for probabilistic considerations within evidence theory (see below). Consider the question: What are the common truth values of the propositions in $N$ ? The set of possible answers to this question (its frame of discernment) is the Boolean cube $B_{n+s}$ and denote by $\theta \in B_{n+s}$ its unknown exact answer. If $\Sigma = \{\xi_1, \ldots, \xi_r\}$ is the available knowledge, then each clause $\xi_i$ restricts the unknown $\theta$ to the set $N(\xi_1) \cap \ldots \cap N(\xi_r)$. If $\xi = \xi_1 \wedge \ldots \wedge \xi_r$, then the knowledge base $\Sigma$ is therefore represented by the subset $N(\xi) = N(\xi_1) \cap \ldots \cap N(\xi_r) \subseteq B_{n+s}$. Similarly, the truth values of the assumptions $a_1, \ldots, a_s$ define a vector $x$ in $B_s$. Note that $B_s$ can be seen as a subset of $B_{n+s}$. If we assume that $x$ is the correct truth vector for the assumptions, then defining

$$\alpha_i = \begin{cases} a_i & \text{if } x_i = 1 \\ \neg a_i & \text{if } x_i = 0 \end{cases} \quad (5)$$

implies that $conj(x) = \alpha_1 \wedge \ldots \wedge \alpha_s$ is true under $x$. So, assuming that $x$ is the correct truth vector for the assumptions permits to restrict the unknown $\theta$ to the subset $N(conj(x))$. Therefore, combining this information with the knowledge base represented by $N(\xi)$ implies that $\theta$ is in $N(conj(x) \wedge \xi)$. Since we know that the probability that $a_i$ is true equals $q_i$ and the assumptions are stochastically independent, the probability that $x$ is the correct vector is

$$P'(x) = \prod_{i=1}^{s} q_i^{x_i}(1-q_i)^{1-x_i}. \quad (6)$$

Remark that the set

$$\Gamma(x) = N(conj(x) \wedge \xi) \quad (7)$$

may very well be empty for some vectors $x$ in $B_s$ because all we know is that if $x$ is the correct vector, then $\theta$ is in $\Gamma(x)$. This implies that if we learn that $\Gamma(x)$ is empty, then $x$ cannot be the correct truth vector of the assumptions. Therefore, the available knowledge $\Sigma$ permits to conclude that the correct vector is in $\Omega = \{x \in B_s : \Gamma(x) \neq \emptyset\}$. This a new piece of information which must be taken into account by *conditioning* the a priori probability $P'$ on the event $\Omega$, resulting in a new probability space $(\Omega, P)$ where $P(x) = P'(x)/P'(\Omega)$. The vectors in $B_s - \Omega$ are called *contradictory configurations* because assuming that $x \in B_s - \Omega$ is the correct configuration leads to a contradiction with the accepted knowledge $\Sigma$ since $N(conj(x)) \cap N(\xi) = \emptyset$, i.e. $conj(x) \wedge \xi$ is inconsistent. These ideas can be placed into the framework of Dempster's multivalued mappings, which makes the link to the theory of evidence (Dempster, 1967). Indeed, $\Gamma$ is a multivalued mapping from $\Omega$ to $B_{n+s}$ and the structure

$$\mathcal{H} = (\Omega, P, \Gamma, B_{n+s}). \quad (8)$$

is called a *hint* (Kohlas, 1990; Kohlas, Monney, 1990, 1992). Now, consider a formula $h \in \mathcal{L}_N$ representing a hypothesis. Notice that $h$ is true if and only if $h$ is true under the interpretation $\theta$, i.e. $\theta \in N(h)$. Therefore, it is interesting to look at the set

$$u(h) = \{x \in \Omega : \Gamma(x) \subseteq N(h)\} \quad (9)$$

because if $x$ is the correct configuration of the assumptions and $x \in u(h)$, then $h$ is necessarily true. Since $x$ is the correct configuration with probability $P(x)$, it is natural to define the *degree of support* of $h$ in the light of the hint $\mathcal{H}$ by

$$sp(h) = P(u(h)). \quad (10)$$

The rest of the paper is concerned with the problem of finding efficient procedures for computing such degrees of support. Now, to compute $sp(h)$, it turns out to be useful to consider a slight generalization of the notion of hint. Indeed, before conditioning, our knowledge is represented by the so-called *quasi-hint*

$$\mathcal{H}' = (B_s, P', \Gamma, B_{n+s}), \quad (11)$$

where this time it is allowed that $\Gamma(x)$ is empty. From a quasi-hint, conditioning always permits to deduce a hint which can be used to compute degrees of support. Note that degrees of support must always be computed with respect to hints and not quasi-hints because contradictory configurations clearly do not contribute to the support or credibility of a hypothesis $h$, although $\Gamma(x) = \emptyset \subseteq N(h)$ for such configurations. A quasi-hint allows for the computation of the so-called *degree of quasi-support* of $h$

$$sp'(h) = P'(u'(h)) \quad (12)$$

where

$$u'(h) = \{x \in B_s : \Gamma(x) \subseteq N(h)\}. \quad (13)$$

If o denotes the inconsistency, then we have the following result.

**Theorem 1** *For any formula* $h \in \mathcal{L}_N$,
$$sp(h) = \frac{sp'(h) - sp'(\circ)}{1 - sp'(\circ)}. \quad (14)$$

The proofs of all theorems in this paper can be found in Kohlas, Monney (1993). This paper also contains several examples. According to this result, in order to compute the degree of support of $h$, we have to compute the degree of quasi-support of $h$ and of o.

## 2 A LOGICAL VIEW OF SUPPORTS OF HYPOTHESES

In this section, we are going to present a method for finding a more economic representation of $u'(h)$ for an arbitrary formula $h \in \mathcal{L}_N$ ($h$ may very well also be the inconsistency o). In fact, for large sets of assumptions $A$, it is very difficult to compute $u'(h)$ explicitly because $B_s$ has $2^s$ elements which must all be tested whether they belong to $u'(h)$. Moreover, in order to compute the degree of support of $h$, we don't really have to compute $u'(h)$, but rather its probability $P'(u'(h))$. Consider the formula

$$\varphi_h = \vee \{conj(x) : x \in u'(h)\} \quad (15)$$

in the sublanguage $\mathcal{L}_A$ of $\mathcal{L}_N$.

**Theorem 2** *Let $h$ be an arbitrary formula in $\mathcal{L}_N$. Then $u'(h) = N(\varphi_h)$.*

Thus we have to compute $P'(N(\varphi_h))$. If we can find a disjunctive normal form representation of $\varphi_h$ which is simpler than (15), say $\varphi_h = f_1 \vee \ldots \vee f_r$, then $sp'(h) = P'(\cup_{i=1}^r N(f_i))$. Computing the probability of a union of events is not too difficult, especially when the $f_i$ are pairwise inconsistent (see section 4). Now, let's look at the problem of finding a simple disjunctive normal form representation of $\varphi_h$. Denote by $\mathcal{C}_A$ the set of all conjunctions of zero, one or more literals in $A^\pm$ (a conjunction with zero literal is the tautology •). We say that a conjunction $a$ in $\mathcal{C}_A$ is a *quasi-support* of $h$ if $a \wedge \xi \models h$. We don't use the term "support" because a conjunction $a$ such that $a \wedge \xi$ is inconsistent is clearly a quasi-support of $h$, but of course we cannot say that it is a support of $h$. Note that ○ is not in $\mathcal{C}_A$ and hence cannot be a quasi-support of $h$. However, each $a_i \wedge \neg a_i, i = 1, \ldots s$ is in $\mathcal{C}_A$ and is in fact a quasi-support of any formula $h$ in $\mathcal{L}_N$.

**Theorem 3** *A conjunction $a \in \mathcal{C}_A$ is a quasi-support of $h \in \mathcal{L}_N$ if and only if $\xi \models h \vee \neg a$.*

A quasi-support of ○ is called a *contradiction*. It is important to note that a contradiction is a quasi-support of any formula $h$ in $\mathcal{L}_N$.

**Theorem 4** *Let $a$ be a conjunction in $\mathcal{C}_A$. The following three statements are equivalent:*

- *$a$ is a contradiction*
- *$\xi \models \neg a$*
- *$a \wedge \xi$ is not satisfiable, i.e. $a \wedge \xi = ○$.*

The following fundamental theorem establishes the link between quasi-supports of $h$ and the logical formula $\varphi_h$.

**Theorem 5** *A conjunction $a \in \mathcal{C}_A$ is an implicant of $\varphi_h$ if and only if $a$ is a quasi-support of $h$. A conjunction $a \in \mathcal{C}_A$ is an implicant of $\varphi_○$ if and only if $a$ is a contradiction.*

A quasi-support of $h$ is called *minimal* if no proper subconjunction of $a$ is also a quasi-support of $h$. A contradiction $a$ is called *minimal* if no proper subconjunction of $a$ is also a contradiction. Be careful, for any formula $h$ in $\mathcal{L}_N$, a minimal contradiction is a quasi-support of $h$ which is minimal within the restricted set of contradictions, but not necessarily within the whole set of quasi-supports of $h$. Therefore, there might exist minimal contradictions which are not minimal quasi-supports of $h$ ! The following corollary follows directly from theorem 5.

**Corollary 1** *A conjunction $a \in \mathcal{C}_A$ is a prime implicant of $\varphi_h$ if and only if $a$ is a minimal quasi-support of $h$. A conjunction $a \in \mathcal{C}_A$ is a prime implicant of $\varphi_○$ if and only if $a$ is a minimal contradiction.*

Again, although each implicant of $\varphi_○$ is also an implicant of $\varphi_h$, note that a prime implicant of $\varphi_○$ is not necessarily a prime implicant of $\varphi_h$.

Now, it is a classical result of logics that if $f_1, \ldots, f_r$ are the prime implicants of $\varphi_h$ then $\varphi_h = f_1 \vee \ldots \vee f_r$ and hence

$$u'(h) = \cup_{i=1}^r N(f_i) \quad (16)$$

by theorem 2. But corollary 1 implies that equation (16) still holds when $f_1, \ldots, f_r$ are the minimal quasi-supports of $h$. Similarly, if $g_1, \ldots, g_s$ are the minimal contradictions, then $\varphi_○ = g_1 \vee \ldots \vee g_s$ and hence

$$u'(○) = \cup_{i=1}^s N(g_i). \quad (17)$$

Formulas (16) and (17) can then be used to compute $sp'(h)$ and $sp'(○)$, which in turn can be used to compute the exact degree of support of $h$ in the light of the available knowledge represented by the hint $\mathcal{H}$. Thus our problem reduces to the determination of all minimal quasi-supports of $h$ and all minimal contradictions. In their "Assumption-Based Truth Maintenance Systems (ATMS)", Reiter, De Kleer (1987) called a contradiction a "no-good" and a quasi-support which is not a contradiction a "label". Let us assume for the moment that $h$ is a *clause*. This assumption will be dropped later. Then, for any quasi-support $a$ of $h$, $\neg a \vee h$ is also a clause which is an implicate of $\xi$. In fact, by adapting theorems of Reiter, De Kleer (1987), it follows that there are even closer links between the minimal quasi-supports of $h$ and the prime implicates of $\xi$ (see below). Let's introduce some notations and conventions. First of all, clauses or conjunctions are sometimes conveniently considered as the sets of the literals they contain. Then for example $x \in c$ means that the literal $x$ appears in $c$, and $c \cap c'$ denotes the set of literals contained both in $c$ and $c'$, but $c \cap c'$ is neither a clause nor a conjunction. A clause or conjunction $f$ is said to *subsume* a clause or a conjunction $g$, if $f \supseteq g$. Furthermore, if $f$ and $g$ are two clauses, let $f - g$ denote the subclause of $f$ which is obtained when all literals present both in $f$ and $g$ are eliminated from $f$. The set of implicates of a set of clauses $\Sigma$ is denoted by $Th(\Sigma)$. For an arbitrary set of clauses $S$ in $\mathcal{L}_N$, the subset of clauses which subsume no other clause of the set, that is, which are minimal within the set, is denoted by $\mu S$. Then the set of prime implicates of $\xi$, i.e. the prime implicates of $\Sigma$, is $PI(\Sigma) = \mu Th(\Sigma)$. Now, let $MQS(\Sigma, h)$ denote the set of minimal quasi-supports of $h$ with respect to $\Sigma$ and $MC(\Sigma)$ the set of minimal contradictions with respect to $\Sigma$. Let $\mathcal{C}'_A$ denote the set of all clauses composed of zero, one or more literals in $A^{\pm}$ (a clause with zero literal is the inconsistency o). Furthermore, note that $\mathcal{C}'_A = \sim \mathcal{C}_A$. The following theorem is a modification of a result of Reiter, De Kleer (1987).

**Theorem 6** *If $h$ is a clause, then*

$$MQS(\Sigma, h) = \sim \mu\{f - h \in \mathcal{C}'_A : f \in PI(\Sigma)\}. \quad (18)$$

*Furthermore,*

$$MC(\Sigma) = \sim \{f \in \mathcal{C}'_A : f \in PI(\Sigma)\} \quad (19)$$

The problem of finding minimal quasi-supports of $h$ and minimal contradictions is thus related to the problem of finding prime implicates of $\xi$. A brute force method based on resolution can be used to find them (see Birkhoff, Bartee (1970)). However, more tightly controlled methods are possible if we adapt techniques pointed out by Inoue (1991 a and b, 1992). They make use of concepts and procedures introduced by Bossu, Siegel (1985) and Siegel (1987). These methods depend on the notions of production field and characteristic clauses (Inoue, 1992). A *production field* $P$ is simply a non empty set of clauses in $\mathcal{L}_N$. In fact, the production field which will interest us most is the set of clauses in $\mathcal{C}'_A$, which will be denoted by $P_A$. A production field is called *stable* if any subsumed clause of a clause in $P$ is also in $P$, i.e. if $c \in P$ and $c' \subseteq c$, then $c' \in P$. Clearly, the empty clause o belongs to any stable production field and $P_A$ is stable. The minimal elements of $Th(\Sigma) \cap P$ are called the *characteristic clauses* of $\Sigma$ with respect to $P$:

$$Carc(\Sigma, P) = \mu(Th(\Sigma) \cap P). \quad (20)$$

Similarly, we can define characteristic clauses for any collection $S$ of formulas in $\mathcal{L}_N$:

$$Carc(S, P) = \mu(Th(S) \cap P). \quad (21)$$

**Theorem 7** *If $P$ is stable production field, then*

$$Carc(S, P) = PI(S) \cap P. \quad (22)$$

Let $f$ be an arbitrary formula in $\mathcal{L}_N$, possibly the inconsistency o. Let $MQS(\Sigma, f)$ denote the minimal quasi-supports of $f$ with respect to $\Sigma$. Given a stable production field $P$, we denote the minimal quasi-supports of $f$ belonging to $\sim P$ by

$$MQS(\Sigma, f, P) = MQS(\Sigma, f) \cap \sim P. \quad (23)$$

In particular, since $MC(\Sigma) = MQS(\Sigma, o)$, we have

$$MQS(\Sigma, o, P) = MC(\Sigma) \cap \sim P. \quad (24)$$

**Theorem 8** *Let $\Sigma$ be a set of clauses in $\mathcal{L}_N$, $f$ an arbitrary formula in $\mathcal{L}_N$ and $P$ a stable production field. Then*

$$MQS(\Sigma, f, P) = \sim Carc(\Sigma \cup \{\neg f\}, P). \quad (25)$$

*Furthermore,*

$$MC(\Sigma) \cap \sim P = \sim Carc(\Sigma, P). \quad (26)$$

Equation (26) can be found in Inoue (1992). In consequence, we have the following result.

**Theorem 9** *If $f$ is an arbitrary formula in $\mathcal{L}_N$, then*

$$MC(\Sigma) = \sim Carc(\Sigma, P_A) \quad (27)$$
$$MQS(\Sigma, f) = \sim Carc(\Sigma \cup \{\neg f\}, P_A). \quad (28)$$

In view of the last theorem, characteristic clauses are of interest to find minimal contradictions and minimal quasi-supports. Note that in theorems 8 and 9, $f$ is an arbitrary formula in $\mathcal{L}_N$, not necessarily a clause $h$ like in theorem 6. In view of these theorems, we are interested in methods to compute characteristic clauses relative to a stable production field $P$, which may be $P_A$ for example. This can be much more feasible than to compute all prime implicates of $\xi$ and apply theorem 6 when $f$ is a clause. Such methods will be presented in the next section.

## 3 COMPUTING CHARACTERISTIC CLAUSES

By theorem 9, it is sufficient to compute the set $Carc(\Sigma, P_A)$ of characteristic clauses with respect to the production field $P_A$ in order to obtain the minimal contradictions with respect to $\Sigma$. Also, by the same theorem, $Carc(\Sigma \cup \{\neg f\}, P_A)$ permits to obtain all minimal quasi-supports of an arbitrary hypothesis $f$ in $\mathcal{L}_N$ (not necessarily a clause). The problem is thus to compute these sets.

Given a set of clauses $\Sigma$, a clause $c$ and stable production field $\mathcal{P}$, Siegel (1987) and Inoue (1992) propose an efficient algorithm called skipped, ordered, linear resolution to compute the set $Prod(\Sigma, c, P)$ consisting of the so-called produced clauses. These sets are fundamental for the computational procedures proposed below. The algorithm for computing $Prod(\Sigma, c, P)$ is therefore the basic computational module. The following theorem of Inoue (1992) shows how characteristic clauses can be computed incrementally.

**Theorem 10** *If $c$ is a clause and $P$ is a stable production field, then*

$$Carc(\emptyset, P) = \{p \lor \neg p : p \in N, p \lor \neg p \in P\}$$

$$Carc(\Sigma \cup \{c\}, P) = \mu(Carc(\Sigma, P) \cup Prod(\Sigma, c, P)).$$

Now, let's apply these results to the computation of minimal contradictions and minimal quasi-supports of hypotheses. The information contained in the set of clauses $\Sigma$ consists of the one side of the relatively stable knowledge base $\Sigma_K$ and on the other hand of the varying facts $\Sigma_F$, which may change from one case to the other. This has to be taken into account, if efficient computational procedures are to be designed. The first task consists in the determination of the minimal contradictions with respect to $\Sigma$. Remind that these minimal contradictions are essentially given by $Carc(\Sigma, P_A)$ (theorem 9). First, the clauses of $\Sigma$ are ordered into a sequence $\xi_1, \xi_2, \ldots, \xi_r$, where it is convenient to take first the $s$ clauses $\xi_1, \ldots, \xi_s$ of $\Sigma_K$. Using theorem 10, the minimal contradictions $Carc(\Sigma_K, P_A)$ relative to the knowledge base $\Sigma_K$ can be computed incrementally. In many practical cases, $Carc(\Sigma_K, P_A)$ will only contain the trivial clauses $a \vee \neg a$ (representing contradictions $a \wedge \neg a$) for all assumptions $a \in A$. Then the knowledge base is in itself consistent, which may even be a requirement for an acceptable knowledge base. Anyway, this computation can be considered as a *compilation* of the knowledge base. The facts will often also arrive sequentially or can in any case be arranged into an arbitrary sequence. Then the set of minimal contradictions $MC(\Sigma)$ can be updated sequentially using theorem 10 to find $MC(\Sigma)$. Indeed, for any new clause $\xi_{t+1}$ arriving, this means essentially determine $Prod(\Sigma, \xi_{t+1}, P_A)$. At this stage, queries can be accepted. Let $h$ be an arbitrary formula in $\mathcal{L}_N$ representing a hypothesis which is queried. By theorem 9, the minimal quasi-supports of $h$ are equal to $\sim Carc(\Sigma \cup \{\neg h\}, P_A)$. If $\neg h$ is a clause, then this set can computed using theorem 10:

$$Carc(\Sigma \cup \{\neg h\}, P_A) = \mu(Carc(\Sigma, P_A) \cup Prod(\Sigma, \neg h, P_A)).$$

If $\neg h$ is not a clause, then $\neg h$ must first be transformed into a conjunctive normal form, preferably a simple one. If $\neg h = h_1 \wedge \ldots \wedge h_t$ and $\mathcal{F} = \{h_1, \ldots, h_t\}$, then $Carc(\Sigma \cup \{\neg h\}, P_A) = Carc(\Sigma \cup \mathcal{F}, P_A)$ and we can again use theorem 10 to compute this set. The approach described so far can be qualified an *interpretative* approach, because any query is treated upon request. In fact it is a combined compiled-interpretative approach, because the minimal contradictions relative to the knowledge base $\Sigma_K$ are computed once for all and the minimal contradictions relative to the information incorporating additional facts are updated as the facts arrive, and the queries are treated as they arrive (only the minimal contradictions with respect to $\Sigma_K$ need not be recomputed). If there are many different queries to treat, a fully *compiled* approach based on theorem 6 might be preferable. These results presuppose the knowledge of $PI(\Sigma)$ to determine the minimal contradictions and the minimal quasi-supports for a *clause* $h$. Note that $PI(\Sigma) = \mu Th(\Sigma) = Carc(\Sigma, P_N)$, that is the set of characteristic clauses of $\Sigma$ with respect to the stable production field $P_N$ consisting of all clauses over $N^\pm$. The latter can be computed using theorem 10. $PI(\Sigma)$ could also be computed using the following theorem of Inoue (1992), which is a slight modification of theorem 10.

**Theorem 11** *If $c$ is a clause, then*

$$\begin{aligned} PI(\emptyset) &= \{p \vee \neg p : p \in N\} \\ PI(\Sigma \cup \{c\}) &= \mu(PI(\Sigma) \cup Prod(PI(\Sigma), c, P_N)). \end{aligned}$$

So, the whole information contained in $\Sigma$ is compiled into and replaced by its prime implicates $PI(\Sigma)$.

Given the decomposition $\Sigma = \Sigma_K + \Sigma_F$, the compilation can again be partitioned into two phases, namely first the compilation of $\Sigma_K$ and then the updating as additional facts arrive. The minimal contradictions are then obtained by equation (19). Any query which is a clause can be immediately treated by using equation (18). The disadvantage of this compiled approach is the large cardinality of the set $PI(\Sigma)$, which already means a big compilation effort for the knowledge base as well as for the updating phase. This presupposes also a convenient data base structure for easy retrieval of elements of $PI(\Sigma)$.

## 4  COMPUTING DEGREES OF SUPPORT

Once the minimal contradictions and the minimal quasi-supports of a formula $h$ in a propositional system have been determined, the degree of support of $h$ can be computed. How exactly this can be done is the subject of this section. Let $h$ be a hypothesis with minimal quasi-supports $f_1, \ldots, f_r$ and minimal contradictions $c_1, \ldots, c_s$. Then $\varphi_h = f_1 \vee \ldots \vee f_r$ and $\varphi_o = c_1 \vee \ldots \vee c_s$ and hence $u'(h) = N(\varphi_h)$ and $u'(\circ) = N(\varphi_o)$ according to theorem 2. Then, using theorem 1, we just have to compute $sp'(h) = P'(N(\varphi_h))$ and $sp'(\circ) = P'(N(\varphi_o))$ to find the degree of support of $h$. The problem is thus to compute $P'(N(\varphi_h))$ and $P'(N(\varphi_o))$. Since these two tasks are of the same kind, it is sufficient to consider techniques to compute $P'(N(\varphi_h))$ only. Because the logical formula $\varphi_h$ defines a Boolean function on $B_s$, $\varphi_h$ can be considered as a binary random variable whose expected value $E(\varphi_h)$ is precisely $sp'(h) = P'(N(\varphi_h))$. We want to point out here that the problem of computing the expected value $E(\varphi)$ of a Boolean function $\varphi$ is a common problem in reliability theory; however the Boolean functions considered there are mainly monotone, which is usually not the case for $\varphi_h$ and $\varphi_o$. But as has been pointed out by Anders (1989), many methods for monotone Boolean functions generalize to arbitrary Boolean functions. We can therefore take advantage of methods from reliability theory to solve our problems. First, remark that the probability of any conjunction $d = \wedge\{l_j : j \in J\}$ of literals $l_j$ in $A^\pm$ can easily be computed. In fact, if we define

$$x_l = \begin{cases} 1 & \text{if } l_j = a_j \\ 0 & \text{if } l_j = \neg a_j, \end{cases} \qquad (29)$$

then

$$P'(N(d)) = \prod_{j \in J} q_j^{x_j}(1-q_j)^{1-x_j}. \qquad (30)$$

On the other hand, we have

$$P'(N(\varphi_h)) = P'(\cup_{i=1}^r N(f_i)) \qquad (31)$$

and the following theorem, called the inclusion exclusion formula, shows how the latter can be computed (Feller, 1964).

**Theorem 12** *We have*

$$P'(\cup_{i=1}^r N(f_i)) = \sum_{k=1}^r (-1)^{k+1} S_k, \qquad (32)$$

*where*

$$S_k = \sum \{P'(\cap_{i \in I} N(f_i)) : I \subseteq \{1,\ldots,r\}, |I| = k\}.$$

Note that the probabilities of the intersections

$$P'(\cap_{i \in I} N(f_i)) = P'(N(\wedge_{i \in I} f_i))$$

are easy to compute because $\wedge_{i \in I} f_i$ is again a conjunction, namely the union of all literals contained in the conjunctions $f_i, i \in I$. It is not excluded that two conjunctions $f_i$ and $f_k$ are contradictory and then $\wedge_{i \in I} f_i = o$, which implies that the corresponding probability is zero. This method to compute $P'(N(\varphi_h))$ is called the *inclusion-exclusion method*. Unfortunately, the sums in (32) contains many terms which makes computations tedious and even impossible for large $r$. Sometimes, the bounds provided by the next theorem allow to stop computations with an acceptable approximation at a relatively small value of $l$. Note however that these bounds are not necessarily monotone in $l$. For monotone Boolean functions there exist better bounds (see for example Kohlas, 1987); for non-monotone Boolean functions less bounds are known today. The following theorem is stated in Barlow, Proschan (1975) and proved in Kohlas, Monney (1993).

**Theorem 13** *If $S_k$ is defined like in theorem 12 and $l$ is a positive integer such that $2l+1 \leq r$, then $sp'(h) \leq S_1$ and*

$$\sum_{k=1}^{2l} (-1)^{k+1} S_k \leq sp'(h) \leq \sum_{k=1}^{2l+1} (-1)^{k+1} S_k. \qquad (33)$$

For hypotheses $h$ with many minimal quasi-supports ($r$ is large), the inclusion-exclusion method may fail to be practicable. In such cases, so-called *disjoint decomposition algorithms* may help. These methods depend on our ability to find a decomposition of $N(\varphi_h)$ into *disjoint* subsets $N(d_k)$ with $d_k \in \mathcal{L}_A$:

$$N(\varphi_h) = \sum_{k=1}^m N(d_k). \qquad (34)$$

If we have such a decomposition, then

$$P'(N(\varphi_h)) = \sum_{k=1}^m P'(N(d_k)). \qquad (35)$$

So, the $N(d_k)$ should be such that

- $P'(N(d_k))$ is easy to compute
- $m$ is as small as possible.

For example, the first requirement is fulfilled when the $d_k$ are conjunctions. The $N(d_k)$ are disjoint if and only if the $d_k$ are pairwise inconsistent, i.e. $d_k \wedge d_l = o$ whenever $k \neq l$. We say that two formulas in $\mathcal{L}_A$ are disjoint if their conjunction is equivalent to o. Furthermore, a disjunction of disjoint formulas will be written as a sum. From the disjunctive normal form representation $\varphi_h = f_1 \vee \ldots \vee f_r$, a representation by *disjoint* terms can easily be obtained (for simplicity, $\wedge$ is written as a product):

$$\begin{aligned}\varphi_h &= f_1 + (\neg f_1) f_2 + (\neg f_1)(\neg f_2) f_3 + \ldots \\ &\quad + (\neg f_1)\ldots(\neg f_{r-1}) f_r.\end{aligned}$$

Note however that these terms are no more conjunctions and their probabilities are not easily computed. In fact the disjunctions $\neg f_j$ must be developed using the distributive law in order to obtain conjunctions. This will in general give far too many terms. A more intelligent way to obtain a disjoint sum representation of $\varphi_h$ has been proposed by Abraham (1979), but for monotone Boolean functions only. But as Anders (1989) noted, the method can easily be adapted to general Boolean functions. The method makes use of the following two results (see Kohlas, Monney (1993)).

**Theorem 14** *Two conjunctions $c'$ and $c''$ are disjoint if and only if there exists a literal $x$ in $c'$ which appears negated $(\neg x)$ in $c''$.*

**Theorem 15** *Suppose that $c'$ and $c''$ are two conjunctions which are not disjoint. Let $X' = \{x_1, \ldots, x_s\}$ be the set of all literals contained in $c'$, but not in $c''$.*

- *If $X'$ is empty, then $c' \vee c'' = c'$.*
- *If $X'$ is not empty, then*
$$\begin{aligned}c' \vee c'' &= c' + (\neg x_1) c'' + x_1 (\neg x_2) c'' + \ldots \\ &\quad + x_1 \ldots x_{s-1}(\neg x_s) c''.\end{aligned}$$

Based on these results the following algorithm can be defined:

For $j = 1$ to $r$
    define $P_{0j} = \{f_j\}$
    for $i = 1$ to $j - 1$
        for all $d_k \in P_{i-1j}$
        if $d_k$ and $f_i$ are disjoint put $d_k$ into $P_{ij}$
        else define $X' = \{x_1, \ldots, x_s\}$ to be the set of all literals in $f_i$ not contained in $d_k$
        if $X'$ is empty drop $d_k$
        else add $(\neg x_1) d_k, x_1(\neg x_2) d_k, \ldots, x_1 \ldots x_{s-1}(\neg x_s) d_k$ to $P_{ij}$.

The next theorem completes the method and proves at the same time its correctness:

**Theorem 16** *Let $P_{j-1j}, j = 1, \ldots, r$ be the sets obtained in the $j$-th loop of the above algorithm. Then*

$$\varphi_h = \sum_{j=1}^r \sum \{d_k \in P_{j-1j}\}.$$

So the above algorithm constructs in fact a disjoint sum representation of $\varphi_h$ by conjunctions. It follows then from theorem 2 that

$$P'(N(\varphi_h)) = \sum_{j=1}^{r} \sum \{P'(N(d_k)) : d_k \in P_{j-1j}\} \quad (36)$$

and the probabilities $P'(N(d_k))$ are easily computed because $d_k$ are all conjunctions. For a small number $r$ of terms in the original disjunctive normal form, this algorithm may well generate more terms than the inclusion-exclusion method; for large $r$ however, its the contrary in general. Abraham's method is not the only method to obtain a disjoint representation of $\varphi_h$. Another method has been proposed by Heidtmann (1989) for monotone Boolean functions. It can also be generalized to arbitrary Boolean functions (Anders, 1989). All these methods have exponential complexity, the problem is NP-hard, as is well known (see for example Ball, 1986).

## 5 CONCLUSION

The mathematical foundations of probabilistic assumption-based reasoning are exposed in the first part of the paper. Then computational aspects are considered and the given methods are alternatives to those traditionally used in the theory of evidence (propagation in Markov trees, see Shenoy, Shafer, 1990). When used in conjunction, these two solution techniques may lead to interesting results, especially in real-world applications when certain parts of the problem are better suited for one method than the other. A computer program implementing the ideas presented in this paper is currently under development.